\documentclass[a4paper,conference]{IEEEtran}
\usepackage{color}
\usepackage{soul}
\usepackage{graphicx}
\usepackage{multirow}
\usepackage{booktabs}
\usepackage{threeparttable}

\ifCLASSINFOpdf
\else
\fi
\begin{document}
%
\title{MMINR: Multi-frame-to-Multi-frame Inference with Noise Resistance for Precipitation Nowcasting with Radar}

\author{\IEEEauthorblockN{Feng Sun \qquad Cong Bai}
\IEEEauthorblockA{College of Computer Science\\and Technology\\
Zhejiang University of Technology\\
310023 Hangzhou, China\\
Email: fsun@zjut.edu.cn, congbai@zjut.edu.cn}
\and
\IEEEauthorblockN{Yi Song}
\IEEEauthorblockA{SPiesat Information\\ Technology Company Limited\\
100195 Beijing, China\\
Email: songyi@piesat.cn\\
}
\and
\IEEEauthorblockN{Jinglin Zhang}
\IEEEauthorblockA{School of Artificial Intelligence\\ Hebei University of Technology\\
300401 Tianjin, China\\
}
}

\maketitle

\begin{abstract}
Precipitation nowcasting based on radar echo maps is essential in meteorological research. Recently, Convolutional RNNs based methods dominate this field, but they cannot be solved by parallel computation resulting in longer inference time. FCN based methods adopt a multi-frame-to-single-frame inference (MSI) strategy to avoid this problem. They feedback into the model again to predict the next time step to get multi-frame nowcasting results in the prediction phase, which will lead to the accumulation of prediction errors. In addition, precipitation noise is a crucial factor contributing to high prediction errors because of its unpredictability. To address this problem, we propose a novel Multi-frame-to-Multi-frame Inference (MMI) model with Noise Resistance (NR) named MMINR. It avoids error accumulation and resists precipitation noise\'s negative effect in parallel computation. NR contains a Noise Dropout Module (NDM) and a Semantic Restore Module (SRM). NDM deliberately dropout noise simple yet efficient, and SRM supplements semantic information of features to alleviate the problem of semantic information mistakenly lost by NDM. Experimental results demonstrate that MMINR can attain competitive scores compared with other SOTAs. The ablation experiments show that the proposed NDM and SRM can solve the aforementioned problems.
\end{abstract}


%
\IEEEpeerreviewmaketitle

\section{Introduction}
Precipitation forecasting, using past rainfall information to predict future rainfall intensity in specific areas, is one of the basic challenges in meteorological research. It is generally used for flood warnings, enhanced airplane flight safety, and so on. In general, this task can be divided into long-term and short-term forecasts. Our paper focuses on the latter, aiming to forecast the future 0-2 hours of precipitation, i.e., precipitation nowcasting.

\begin{figure}[!t]
  \flushleft
  \includegraphics[width=0.49\textwidth]{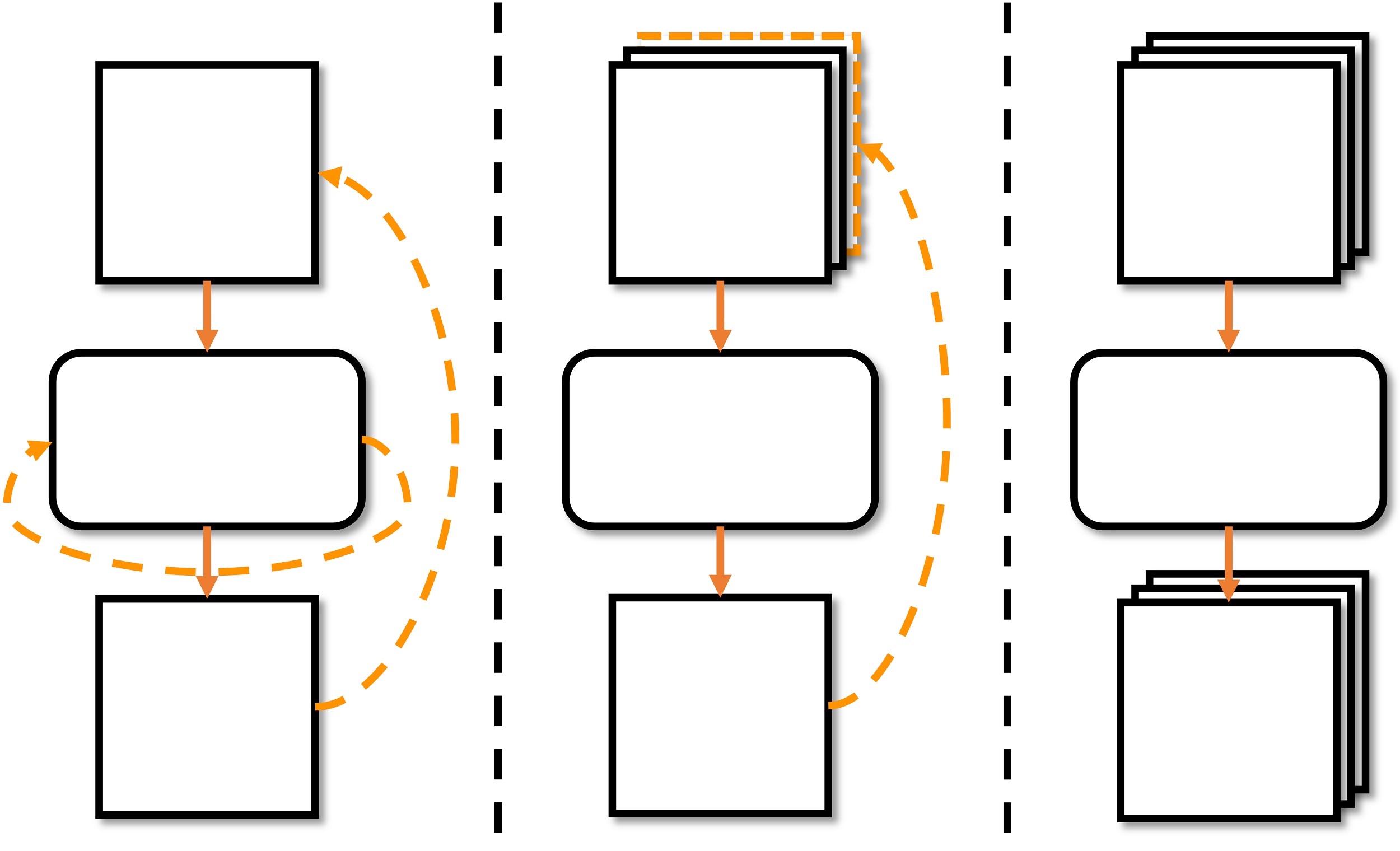}
  \put(-134, 71){\normalsize FCNs}
  \put(-238, 71){\normalsize ConvRNNs}
  \put(-48, 71){\normalsize MMINR}
  \put(-95, 50){\small feed}
  \put(-95, 42){\small back}
  \put(-200, 50){\small recurrent}
  \put(-134, 124){\small input:}
  \put(-139, 114){\small sequence}
  \put(-226, 124){\small input:}
  \put(-226, 114){\small frame}
  \put(-42, 124){\small input:}
  \put(-47, 114){\small sequence}
  \put(-44, 28){\small output:}
  \put(-47, 18){\small sequence}
  \put(-228, 28){\small output:}
  \put(-226, 18){\small frame}
  \put(-135, 28){\small output:}
  \put(-133, 18){\small frame}
  \put(-220, -5){\small (a)}
  \put(-127, -5){\small (b)}
  \put(-35, -5){\small (c)}
  \caption{Different data flow in (a) ConvRNN based model, (b) FCN based model and (c) The proposed MMINR.}
  \label{fig:differentModel}
\end{figure}
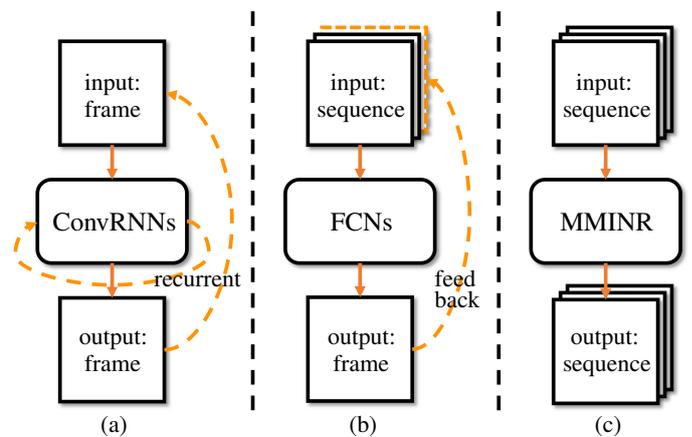

\textsl{Numerical Weather Prediction} (NWP)~\cite{sun2014use} is a common method in weather forecasting, which uses large amounts of weather data and supercomputers to simulate the state of the natural atmospheric system. Therefore, it requires expensive computational resources and much complex weather data. At this point, the radar extrapolation-based method is another excellent alternative. It only needs to use radar echo maps to predict, so from another perspective, the precipitation prediction task can also be seen as a computer vision task. Hence, radar echo maps based precipitation nowcasting task can be regarded as a sequence-to-sequence prediction task, in which deep learning based methods have achieved excellent results in computer vision domain.

\begin{figure}[!t]
  \flushleft
  \includegraphics[width=0.49\textwidth]{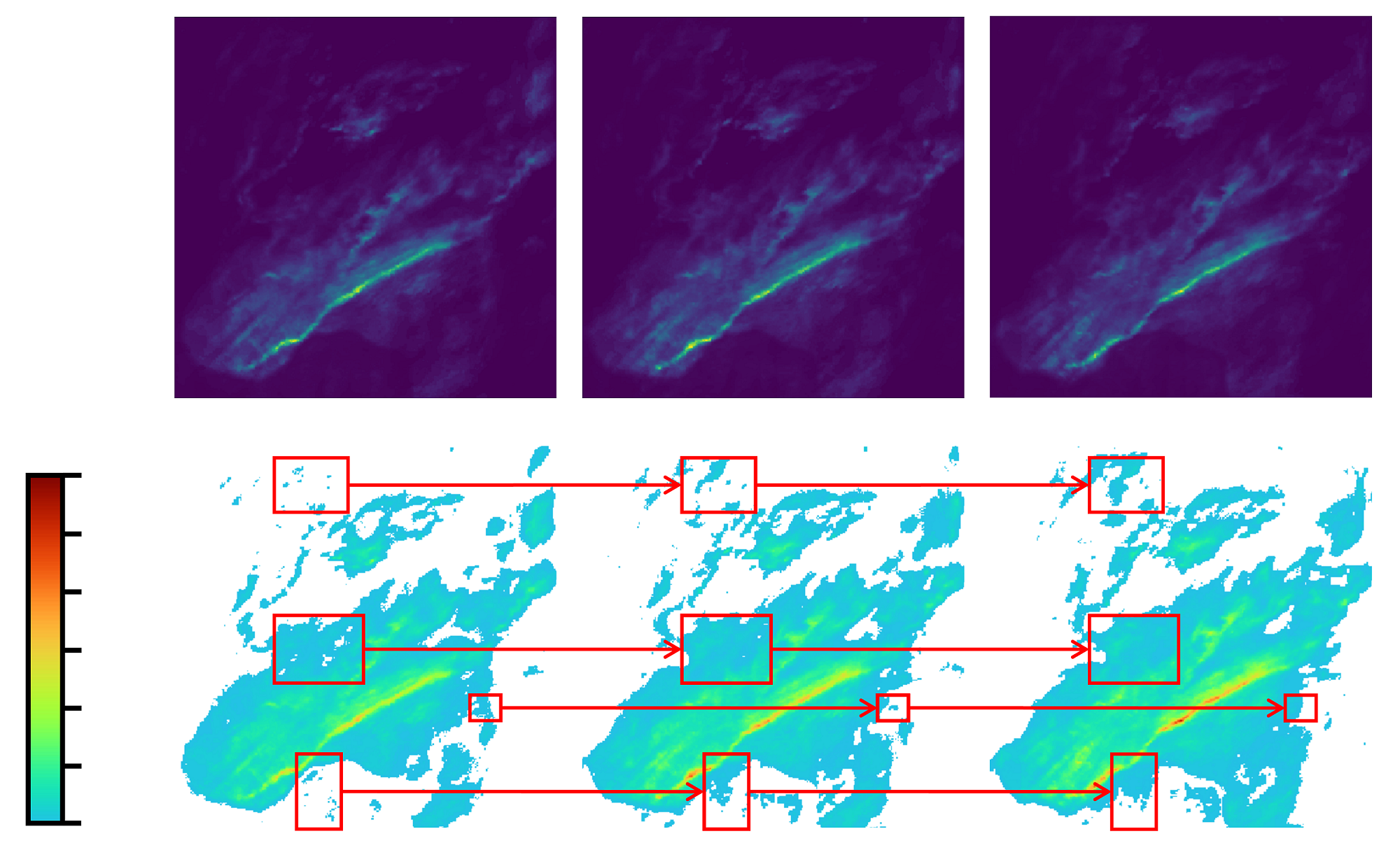}
  \put(-192, 77){\footnotesize (a)}
  \put(-118, 77){\footnotesize (b)}
  \put(-44, 77){\footnotesize (c)}
  \put(-192, -3){\footnotesize (d)}
  \put(-118, -3){\footnotesize (e)}
  \put(-44, -3){\footnotesize (f)}
  \put(-196, 155){\footnotesize $T-1$}
  \put(-116, 155){\footnotesize $T$}
  \put(-51, 155){\footnotesize $T+1$}
  \put(-237, 5){\scriptsize 0.5}
  \put(-235, 15){\scriptsize 5}
  \put(-237, 26){\scriptsize 10}
  \put(-237, 36){\scriptsize 15}
  \put(-237, 47){\scriptsize 20}
  \put(-237, 57){\scriptsize 25}
  \put(-237, 68){\scriptsize 30}
  \put(-227, 69){\rotatebox{270}{\scriptsize Precipitation (mm/h)}}
  \caption{The precipitation noise in radar echo maps. (a), (b) and (c) are continuous raw radar echo maps in a sequence. (d), (e) and (f) are coloring maps. The red-bound area is precipitation noise, which usually vanishes, arises, or deforms suddenly between frames. It is meaningless to pay much attention to it. In short, these regions are unpredictable.}
  \label{fig:precipitationNoise}
\end{figure}

Currently, most of deep learning-based methods in precipitation nowcasting can roughly be divided into \textsl{Convolutional Recurrent Neural Network} (ConvRNN)~\cite{shi2015convolutional} based methods and \textsl{Fully Convolutional Network} (FCN)~\cite{long2015fully} based methods. ConvRNNs can extract both temporal and spatial features simultaneously, which achieves outstanding performance on prediction tasks. However, it calculates feature maps with the step of time, as shown in the (a) of Fig.~\ref{fig:differentModel}. This means that it cannot be solved by parallel computation and needs a longer calculation time. In addition, ConvRNN based methods have a relatively complex unit structure, resulting in difficulty for training and being prone to gradient explosion when the model is deeper~\cite{wang2018predrnn++}. In contrast, many FCN based methods can be solved by parallel computation in the training phase with multi-frame as input and single frame as output. In the prediction phase, the single-frame resulting from the previous time step will feedback into the model as input to predict the following step result, as shown in the (b) of Fig.~\ref{fig:differentModel}. Due to this strategy, the prediction errors will be accumulated in the recurrent process, resulting in the performance of FCN based methods worse than ConvRNNs. Furthermore, some unpredictable regions, known as precipitation noise, as shown in Fig.~\ref{fig:precipitationNoise}, will increase the accumulation of errors. The interval between two consecutive radar frames is often up to several minutes, and with the extreme complexity of the atmospheric system, the sudden vanish or appearance of these noises is almost unpredictable. But these noises attracts much attention from the model and degrades the model performance. Hence, we expect a model that can satisfy the following properties to address the above issues:

\begin{enumerate}
\item[(a)] The model can be solved by parallel computation.
\item[(b)] The model has a simple architecture.
\item[(c)] The model can avoid the accumulation of prediction errors.
\item[(d)] The model can deal with precipitation noise.
\end{enumerate}

This paper proposes a \textsl{Multi-frame-to-Multi-frame Inference model with Noise Resistance} for precipitation nowcasting based on radar echo maps, named MMINR. Since MMINR does not contain any recursive structure and has a simple architecture, it satisfies requirements {\bf{(a)}} and {\bf{(b)}}. Instead of following the circular or single-frame inference model of previous FCN based models, we use multi-frame input to generate multi-frame output at once by adding the output channels in the last convolution layer to reduce the accumulation of prediction errors. So it satisfies the condition {\bf{(c)}}. \textsl{Noise Resistance} (NR) contains \textsl{Noise Dropout Module} (NDM) and \textsl{Semantic Restore Module} (SRM).  NDM reduces noise propagation to the deeper layers of the network by reducing the number of channels layer by layer. To compensate for the semantic information lost by NDM while losing noise, SRM is proposed to enhance the semantic features fusion in the decoder. So MMINR can meet the requirement {\bf{(d)}}.

To summarize, the contributions of our work are as follows:

\begin{itemize}
\item A novel multi-frame-to-multi-frame inference model named MMINR is proposed for precipitation nowcasting. It has multi-output channels in the last convolution layer. Compared with other FCN based methods, it significantly reduces the accumulation of prediction errors.
\item A new Noise Resistance (NR) strategy is proposed that contains noise dropout module (NDM) in the encoder and semantic restore module (SRM) in the decoder. NDM discards precipitation noise by reducing the features channels layer by layer, while SRM enhances the mining of semantic information by fusion the previous and current time step semantic information.
\item Experimental results on ablation study and comparison with ConvRNN based SOTAs and FCN based SOTAs show that MMINR attains competitive scores. The proposed NDM and SRM alleviate the impact of precipitation noise and enhance the extraction of semantic information.
\end{itemize}

\section{Related Works}

\begin{figure*}[!t]
\centering
\includegraphics[width=0.95\textwidth]{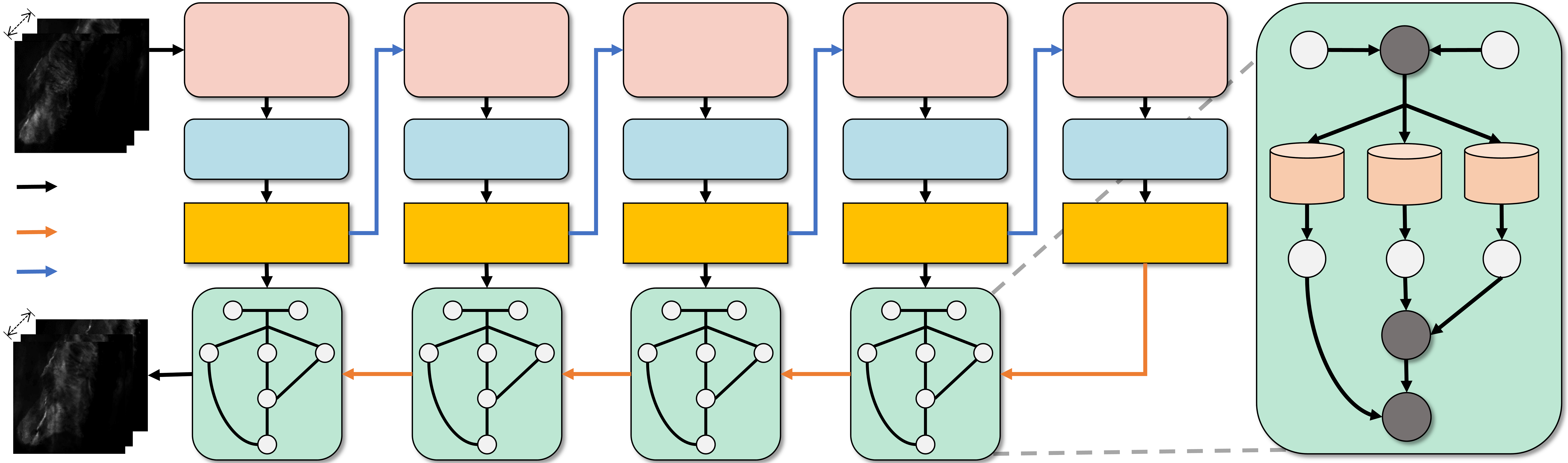}
\put(-492, 140){\tiny $n_1$}
\put(-492, 46){\tiny $n_2$}
\put(-470, 85){\scriptsize feature flow}
\put(-470, 71){\scriptsize upsample}
\put(-470, 57){\scriptsize downsample}
\put(-418, 147){\scriptsize Stage 1}
\put(-415, 126){\scriptsize NDM}
\put(-418, 96){\scriptsize CBAM}
\put(-402, 8){\scriptsize SRM}
\put(-423, 73){\scriptsize $feature_1$}
\put(-431, 66){\tiny $288\times288\times256$}
\put(-349, 147){\scriptsize Stage 2}
\put(-346, 126){\scriptsize NDM}
\put(-349, 96){\scriptsize CBAM}
\put(-333, 8){\scriptsize SRM}
\put(-354, 73){\scriptsize $feature_2$}
\put(-362, 66){\tiny $144\times144\times128$}
\put(-280, 147){\scriptsize Stage 3}
\put(-277, 126){\scriptsize NDM}
\put(-280, 96){\scriptsize CBAM}
\put(-265, 8){\scriptsize SRM}
\put(-285, 73){\scriptsize $feature_3$}
\put(-288, 66){\tiny $72\times72\times64$}
\put(-212, 147){\scriptsize Stage 4}
\put(-209, 126){\scriptsize NDM}
\put(-212, 96){\scriptsize CBAM}
\put(-196, 8){\scriptsize SRM}
\put(-217, 73){\scriptsize $feature_4$}
\put(-220, 66){\tiny $36\times36\times32$}
\put(-143, 147){\scriptsize Stage 5}
\put(-140, 126){\scriptsize NDM}
\put(-143, 96){\scriptsize CBAM}
\put(-148, 73){\scriptsize $feature_5$}
\put(-151, 66){\tiny $18\times18\times32$}
\put(-54, 127){\scriptsize $C$}
\put(-62, 138){\scriptsize connect}
\put(-84, 138){\scriptsize $f_i$}
\put(-29, 138){\scriptsize $f^{'}_{i+1}$}
\put(-22, 52){\tiny Weaken}
\put(-22, 46){\tiny Matrix}
\put(-80, 52){\tiny Boost}
\put(-80, 46){\tiny Matrix}
\put(-44, 66){\tiny Fusion}
\put(-44, 60){\tiny Feature}
\put(-92, 86){\scriptsize $Conv_1$}
\put(-62, 86){\scriptsize $Conv_2$}
\put(-32, 86){\scriptsize $Conv_3$}
\put(-18, 73){\normalsize $\sigma$}
\put(-55, 37){\normalsize $\times$}
\put(-42, 32){\tiny Element-wise}
\put(-42, 27){\tiny Multiplication}
\put(-54, 12){\normalsize $+$}
\put(-40, 12){\tiny Element-wise}
\put(-40, 7){\tiny Summation}
\caption{The flowchart of MMINR.}
\label{fig:architecture}
\end{figure*}
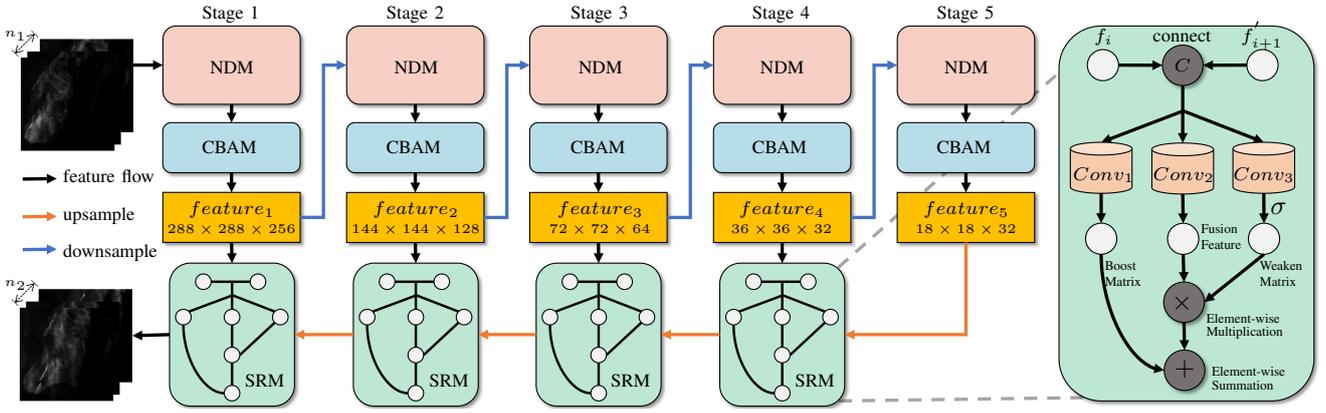

\subsection{ConvRNN based models for precipitation nowcasting}

In precipitation nowcasting, ConvRNN based methods achieve an excellent effect. ConvLSTM~\cite{shi2015convolutional} is the first ConvRNN based method for precipitation nowcasting. It combines convolutional operation with LSTM to extract spatial and temporal information. PredRNN~\cite{wang2017predrnn} enhances the spatial information extraction ability of the model through extra spatial memory units. PredRNN++~\cite{wang2018predrnn++} proposed a Causal LSTM unit and gradient highway unit to alleviate the gradient propagation difficulties. MIM~\cite{wang2019memory} decomposes the spatial-temporal information into stationary and non-stationary features and then uses the corresponding unit to predict the features. Although the ConvRNN based methods are designed for spatiotemporal forecasting problem, it suffers from an unavoidable problem that they cannot be computed in parallel, resulting in more extended calculation and training time.

\subsection{FCN based models for precipitation nowcasting}

FCN based models can be solved by parallel computation but lack the sensitivity to temporal information. Many FCN based methods adopt a recurrent strategy. It inputs multi-frame to the model and outputs single frame as prediction result, then it feeds the result into the model again to predict the next frame result and finally multi-frame prediction results can be got. RainNet~\cite{ayzel2020rainnet} follows the model of Unet~\cite{ronneberger2015u}, which adopts multi-frame-to-single-frame inference in the training phase and recurrent strategy in the testing phase to achieve multiple frames prediction as output. SmaAt-UNet~\cite{trebing2021smaat} aims at lightweight. It reduces the number of convolutions in Unet and uses the attention module to compensate for missing semantic information, drastically reducing the model size. However, SmaAt-UNet only predicts one future frame and does not have the capability of continuous prediction. RDCNN~\cite{shi2018method} contains a recurrent structure based on CNN that is different from LSTM. It auto generates convolution kernel from input data. But RDCNN still does not avoid the accumulation of prediction errors. Therefore, recurrent prediction leading to prediction error accumulation is an urgent problem to be solved.

\section{Methodology}

\subsection{Problem definition}

From the perspective of computer vision, we define the precipitation nowcasting problem as a sequence-to-sequence problem. In this task, the input data and prediction result are the sequences of radar echo maps. $X$ and $\hat{Y}$ denote the input data and the prediction result respectively. And $\hat{Y}$ has the same size as $X$. $X=\{ x_1,x_2,...,x_N \}$ is a collection of $N$ radar echo maps $x_i$, each with the same size.  $\hat{Y}=\{\hat{y_1},\hat{y_2},...,\hat{y_M}\}$ is a collection of $M$ prediction frames $\hat{y_i}$. So the prediction process can be defined as the following:
\begin{equation}
\hat{Y} = \Gamma(X).
\end{equation}
where $\Gamma$ represents the precipitation nowcasting model.

\subsection{Multi-frame-to-Multi-frame Inference (MMI)}
ConvRNN based models and FCN based models contain different cyclic structure in precipitation nowcasting, as shown in Fig.~\ref{fig:differentModel}. FCN based model receives multiple frames as input and outputs a single frame as a prediction result. Then it feed the prediction result back into the model. Such process can be defined as:
\begin{equation}
\hat{y}_{n+1} = F(x_0, x_1, ..., x_n),
\end{equation}
\begin{equation}
\hat{y}_{n+2} = F(x_1, ..., x_n, \hat{y}_{n+1}).
\end{equation}
where $x_i$ and $\hat{y}_j$ represent the i-th input frame and the j-th prediction result, respectively. $F$ means FCN based model. Many FCN based models will predict a single frame in the training phase, parallelize computation, and use the recurrent prediction method mentioned above in the testing phase.

ConvRNN based models receive only one frame as input at each time step and uses memory units to transfer features between time steps, as shown in (a) of Fig.~\ref{fig:differentModel}. Both FCNs or ConvRNN based methods have the same characteristic. In the testing phase, they take the output from the previous time step as input of the next step. Due to the output at the previous time step must contain the prediction error, the next time step will accumulate error in the recurrent process.

\setlength{\tabcolsep}{2.4mm}{
\begin{table*}[!t]
\caption{Quantitative evaluation results.\label{tab:quantitativeResults}}
\centering
\begin{threeparttable}
\begin{tabular}{l|c|c|c|c|c|c|c|c|c|c}
\hline
\multicolumn{11}{c}{9 frames to 1 frame (9-to-1)} \\
\hline
\multirow{2}{*}{Method} & \multicolumn{4}{c|}{CSI/frame $\uparrow$} & \multicolumn{4}{c|}{HSS/frame $\uparrow$} & \multirow{2}{*}{B-MSE $\downarrow$} & \multirow{2}{*}{B-MAE $\downarrow$}\\
\cline{2-9}
& r\tnote{1} $\ge$ 0.5 & r $\ge$ 2 & r $\ge$ 5 & r $\ge$ 10 & r $\ge$ 0.5 & r $\ge$ 2 & r $\ge$ 5 & r $\ge$ 10 & & \\
\hline
SmaAt-UNet~\cite{trebing2021smaat} (MSI) & \bf0.8321 & 0.6083 & 0.3151 & 0.1290 & \bf0.4306 & 0.3558 & 0.2056 & 0.0915 & 1.8954 & 0.5359 \\
MMINR-MSI & 0.8247 & \bf0.6254& \bf0.3399 & \bf0.1371 & 0.4217 & \bf0.3628 & \bf0.2183 & \bf0.0963 & \bf1.8286 & \bf0.5031 \\
\hline
\multicolumn{11}{c}{9 frames to 9 frames (9-to-9)} \\
\hline
RainNet~\cite{ayzel2020rainnet} (MSI + Recurrent) & 0.6308 & 0.3471 & 0.1268 & 0.0362 & 0.3094 & 0.2093 & 0.0881 & 0.0274 & 6.3938 & 1.1728 \\
MMINR-MSI+Recurrent\tnote{2} & 0.6235 & 0.3560 & 0.1321 & 0.0424 & 0.3068 & 0.2169 & 0.0922 & 0.0323 & 5.4057 & 1.1401 \\
\hline
ConvLSTM{}~\cite{shi2015convolutional} & 0.6784 & 0.3763 & 0.1279 & 0.0242 & 0.3442 & 0.2311 & 0.0900 & 0.0192 & 4.8311 & 0.9695 \\
PredRNN~\cite{wang2017predrnn} & 0.6790 & 0.3686 & 0.1265 & 0.0254 & 0.3449 & 0.2271 & 0.0890 & 0.0200 & 4.9645 & 0.9738 \\
MIM~\cite{wang2019memory} & 0.6845 & 0.3730 & 0.1261 & 0.0248 & 0.3483 & 0.2297 & 0.0882 & 0.0194 & 4.8804 & 0.9611 \\
PredRNN++~\cite{wang2018predrnn++} & 0.6817 & 0.3693 & 0.1252 & 0.0274 & 0.3463 & 0.2277 & 0.0876 & 0.0212 & 4.9265 & 0.9710 \\
PFST-LSTM~\cite{luo2020pfst} & \bf0.6858 & 0.3736 & 0.1326 & 0.0260 & \bf0.3498 & 0.2303 & 0.0926 & 0.0203 & \bf4.7077 & \bf0.9453 \\
\hline
MMINR & 0.6769 & \bf0.3778 & \bf0.1405 & \bf0.0412 & 0.3439 & \bf0.2312 & \bf0.0970 & \bf0.0310 & 4.9284 & 0.9750 \\
MMINR-SRM & 0.6721 & 0.3636 & 0.1294 & 0.0364 & 0.3396 & 0.2228 & 0.0891 & 0.0270 & 4.8176 & 0.9769 \\
MMINR-SRM-NDM & 0.6686 & 0.3620 & 0.1209 & 0.0253 & 0.3361 & 0.2228 & 0.0824 & 0.0191 & 4.9516 & 0.9937 \\
\hline
\end{tabular}
\begin{tablenotes}
  \scriptsize
  \item[1] r represents the intensity of rainfall. The unit is mm/h.
  \item[2] This method is only used for fair comparison with RainNet and is not compared with other SOTAs.
\end{tablenotes}
\end{threeparttable}
\end{table*}
}

\setlength{\tabcolsep}{3mm}{
\begin{table}
\caption{rain rate statistic of the KNMI.}\label{tab:rateStatistics}
\centering
\begin{tabular}{l|l|l}
\hline
Rain Rate (mm/h) & Proportion & Rainfall Level \\
\hline
0 \ \ \ $\le$ \ \ x \ \ $\le$ 0.5 & 63.7534 \% & No \\
0.5 \ $\le$ \ \ x \ \ $\le$ 2 & 25.6244 \% & Ligh \\
2 \ \ \ $\le$ \ \ x \ \ $\le$ 5 & {\color{white}0}8.7806 \% & Ligh to moderate \\
5 \ \ \ $\le$ \ \ x \ \ $\le$ 10 & {\color{white}0}1.5652 \% & Moderate \\
10 \ \ $\le$ \ \ x & {\color{white}0}0.2764 \% & Moderate to heavy \\
\hline
\end{tabular}
\end{table}}

As shown in (c) of the Fig.~\ref{fig:differentModel}, we propose the Multi-frame-to-Multi-frame Inference (MMI) model to avoid error accumulation. It takes a sequence (i.e., multiple frames) as input and directly outputs a sequence. The total process can be written as:
\begin{equation}
\hat{y}_{n+1}, \hat{y}_{n+2}, ..., \hat{y}_{n+m} = \phi(x_0, x_1, ..., x_n).
\end{equation}
in which $\phi$ means MMI. $n$ and $m$ represent the number of input and output frames. In general, the last layer in other FCN based model are a single convolution operation with one output channel. But in our method, the last convolution layer has $m$ output channels.

\subsection{Noise Resistance}
Due to the time interval between two continuous radar frames usually up to several minutes, some rainfall areas vanish, arise, or deform suddenly between frames. These areas are almost unpredictable and should be seen as precipitation noises. Those noises will be transmitted to deeper layers, resulting in the model performance degradation in the classical FCN based model, e.g., RainNet~\cite{ayzel2020rainnet}. To address this problem, we design a novel \textsl{Noise Resistance} (NR) strategy, which is composed of NDM in encoder and SRM in decoder.

\subsubsection{NDM}
To dropout the noise in the echo map, five stages of processing are constructed, in which each stage contains a NDM. NDM uses convolutional operations to extract features from the input data. Unlike other typical FCN based models, our approach decreases the number of feature channels with the deeper network. More specifically, the first NDM extracts features from the original input radar maps and generates shallow features with 256 channels. The other NDMs further mine the input features and reduce the number of features channels by 1/2 to drop precipitation noise.

NDM drops part of semantic features along with the noise. To alleviate the information loss problem, we add attention mechanism block \textsl{Convolutional Block Attention Module} (CBAM)~\cite{woo2018cbam} after each NDM to further mine the semantic features. The process can be defined as:
\begin{equation}
f_i=CBAM(NDM(X)), i\in\{1\}
\end{equation}
\begin{equation}
f_i=CBAM(NDM(Down(f_{i-1}))), i\in\{2,3,4,5\}
\end{equation}
where $f_i$ represent the feature from the i-th stage and $Down$ means the downsample.

\subsubsection{SRM}
Despite CBAM enhancing the model's semantic information mining ability, the semantic features dropped in the previous stage still cannot be transferred to the current stage. We designed a \textsl{Semantic Restore Module} (SRM) to retrieve the lost semantic information to deal with this problem. The structure of SRM is shown on the right in Fig.~\ref{fig:architecture}.

SRM receives two input data, features of the current stage, record as $f_i$, and features of the latter stage after upsampling, $f^{'}_{i+1}$. Due to $f_i$ containing the semantic information lost by $f^{'}_{i+1}$, SRM adaptively captures semantic features from $f_i$ to supplement $f^{'}_{i+1}$. Firstly, they are combined by the connect operation. Then three features are extracted by three different convolutional modules, recorded as boost matrix, fusion feature, and weaken matrix. All values in weaken matrix are between 0 and 1 by a sigmoid function. It is multiplied with the fusion feature to reduce some of the feature values that are considered unwanted. The boost matrix is added with the weakened fusion feature to strengthen the values that need to be enhanced.

\setlength{\tabcolsep}{3mm}{
\begin{table}
\caption{rain rate statistic of the KNMI.}\label{tab:SOTAsClass}
\centering
\begin{threeparttable}
\begin{tabular}{l|l|l|l}
\hline
Category & Methods & Traning & Testing \\
\hline
\multirow{3}{*}{ConvRNNs} & ConvLSTM, PredRNN, & \multirow{3}{*}{Recurrent\tnote{1}} & \multirow{3}{*}{Recurrent} \\
& MIM, PredRNN++, & & \\
& PFST-LSTM & & \\
\hline
\multirow{2}{*}{FCNs}  & \multirow{2}{*}{RainNet~\cite{ayzel2020rainnet}} & \multirow{2}{*}{MSI\tnote{2}} & MSI + \\
& & & Recurrent \\
\hline
FCNs & SmaAt-UNet~\cite{trebing2021smaat} & MSI & MSI \\
\hline
FCNs & Ours & MMI & MMI \\
\hline
\end{tabular}
\begin{tablenotes}
  \scriptsize
  \item[1] The cyclic structure that feeds the predicted frame back in again.
  \item[2] Multi-frame-to-Single-frame Inference (MSI).
\end{tablenotes}
\end{threeparttable}
\end{table}}

\section{Experiments}

\subsection{Database}

\begin{figure*}[t]
\centering
\includegraphics[width=0.93\textwidth]{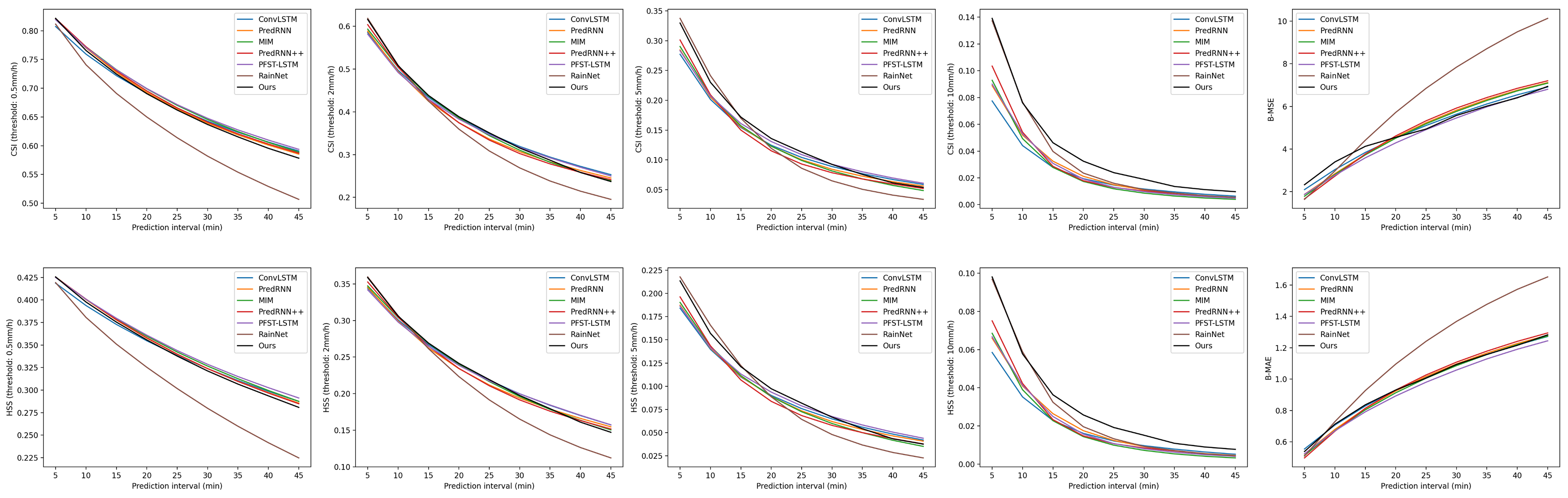}
\caption{Prediction comparison with different prediction interval.}
\label{fig:curve}
\end{figure*}

Our model is trained and tested in a precipitation nowcasting benchmark database proposed by~\textsl{Koninklijk Nederlands Meteorologisch Instituut} (KNMI). This database has 420,000 radar echo maps of Netherlands in 5-min intervals and~\cite{trebing2021smaat} converts the radar maps to rainfall maps. It contains 5734 frame sequences in the training set, 1557 sequences in the test set. We chose randomly 5000 sequences from the default training set as the training set and the rest of 734 as the validation set. Every sequence includes 18 frames, with the size of 288$\times$288 pixels. As shown in Table~\ref{tab:rateStatistics}, we divide the rainfall maps by four thresholds, 0.5 mm/h, 2 mm/h, 5 mm/h, 10 mm/h. They represent light rain, light to moderate rain, moderate rain, and moderate to heavy rain respectively. We train all methods in our experiments on training set. The results of all methods are obtained on testing set. In order to avoid sudden rainfall intensity that harms the training process, we assign 19 mm/h to any rainfall intensity bigger than 19mm/h. For calculation convenience, we deal data with formula as

\begin{equation}
y = \frac{ln(e, x+1)}{1.5}-1
\end{equation}

$x$ represents raw precipitation data, $y$ represents the processed data. This formula keeps the data between -1 and 1 and allocates a suitable interval for high rainfall intensity.

\subsection{Evaluation Criteria}
For comprehensive evaluation, we adopt two binary metrics and two non-binary metrics. Binary metrics binarize the prediction results and focus on the accuracy of prediction results on different rainfall intensity predictions. Non-binary metrics concentrate on the similarity of the prediction results and GT. \textsl{Critical Success Index} (CSI) and \textsl{Heidke Skill Score} (HSS)~\cite{hogan2010equitability} are binary metrics. CSI measures the fraction of observed and/or forecast events that are correctly predicted and HSS measures the ratio of the odds of making a hit to the odds of making a false alarm. All binary evaluation metrics except correlation are based on true negative (TN), false positive (FP), false negative (FN), and true positive (TP). For binary rainfall map, we define 0.5mm/h, 2mm/h, 5mm/h, 10mm/h as the specific threshold. Due to the extreme imbalance of the precipitation dataset,~\cite{shi2017deep} proposes \textsl{Balance-MSE} (B-MSE) and \textsl{Balance-MAE} (B-MAE) which assign weights to rainfall areas according to rainfall intensities to mitigate the negative impact of data imbalance.

\subsection{Implementation Details}
PyTorch implements MMINR with a station equipped with NVIDIA GeForce RTX 2080Ti GPU. We set 0.0001 as the initial learning rate and use Adam optimizer for stochastic gradient descent method. We use a mini-batch of 16 sequences. B-MAE is used as the validation loss function. When the validation loss is no longer decreasing during the training phase, the model with the smallest validation loss is selected as model well trained for prediction.

\subsection{Quantitative Comparisons with SOTAs}

For a comprehensive comparison between MMINR and other SOTAs, as shown in Tab.~\ref{tab:SOTAsClass}, we classify all methods into four categories based on the training and testing strategies used in the original papers. For fair comparison and to explore the model performance, we also provide variants of our models for comparison with other FCN based methods.

\subsubsection{Comparisons with FCN based models}
For fair comparison with the representative FCN based model RainNet, we provide a variant version of MMINR that adopts multi-frame-to-single-frame inference in the training phase and recurrent strategy in the testing phase to achieve multiple frames prediction. This is the same with RainNet, and we name it as MMINR-MSI+Recurrent. From Tab.~\ref{tab:quantitativeResults}, we can see that MMINR-MSI+Recurrent outperforms RainNet except in light rain prediction. The reason is that MMINR loses a part of light rain information as light rain has largest proportion in the database as shown in Table~\ref{tab:rateStatistics}. After adopting the MMI structure, MMINR substantially outperforms RainNet in all metrics. This is because our method directly generates a sequence to avoid the accumulation of prediction errors. Besides, we compare MMINR with SmaAt-UNet at nine frames inferring a single frame mode. The data show that MMINR also outperforms SmaAt-UNet.

\begin{figure}[t]
\centering
\includegraphics[width=0.43\textwidth]{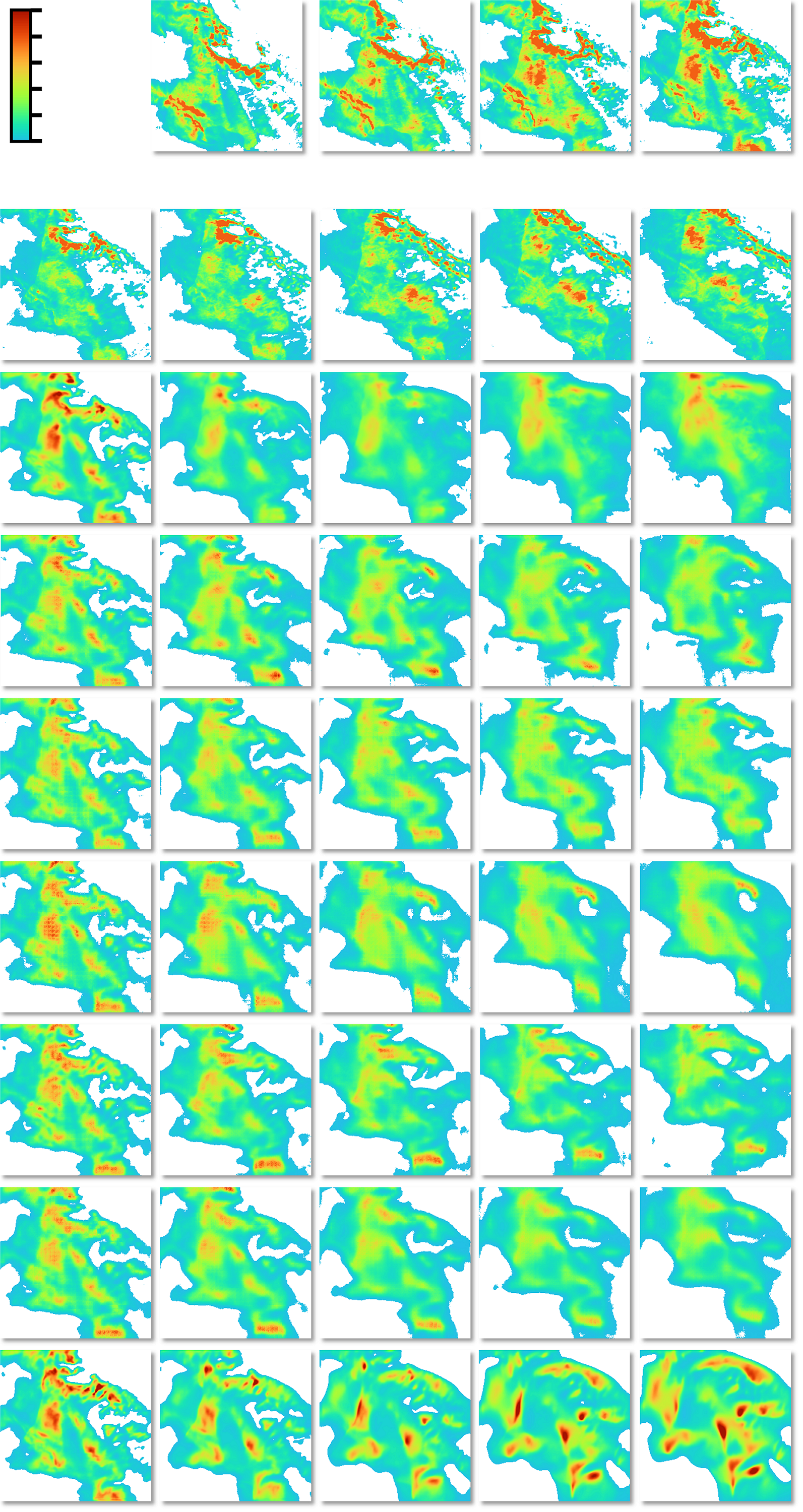}
\put(-172, 423){\scriptsize T-30 min}
\put(-127, 423){\scriptsize T-20 min}
\put(-82, 423){\scriptsize T-10 min}
\put(-35, 423){\scriptsize T min}
\put(-214, 365){\scriptsize T+5 min}
\put(-172, 365){\scriptsize T+15 min}
\put(-127, 365){\scriptsize T+25 min}
\put(-84, 365){\scriptsize T+35 min}
\put(-40, 365){\scriptsize T+45 min}
\put(-210, 414){\tiny 25}
\put(-210, 407){\tiny 20}
\put(-210, 400){\tiny 15}
\put(-210, 392){\tiny 10}
\put(-208, 385){\tiny 5}
\put(-210, 378){\tiny 0.5}
\put(-200, 376){\rotatebox{90}{\tiny Precipitation (mm/h)}}
\put(-1, 388){\rotatebox{90}{\scriptsize Input}}
\put(-1, 332){\rotatebox{90}{\scriptsize GT}}
\put(-1, 287){\rotatebox{90}{\scriptsize Ours}}
\put(-1, 231){\rotatebox{90}{\scriptsize ConvLSTM}}
\put(-1, 190){\rotatebox{90}{\scriptsize PredRNN}}
\put(-1, 149){\rotatebox{90}{\scriptsize MIM}}
\put(-1, 96){\rotatebox{90}{\scriptsize PredRNN++}}
\put(-1, 50){\rotatebox{90}{\scriptsize PFST-LSTM}}
\put(-1, 11){\rotatebox{90}{\scriptsize RainNet}}
\caption{Visual comparisons of the proposed MMINR and the state-of-the-art methods.}
\label{fig:visual}
\end{figure}
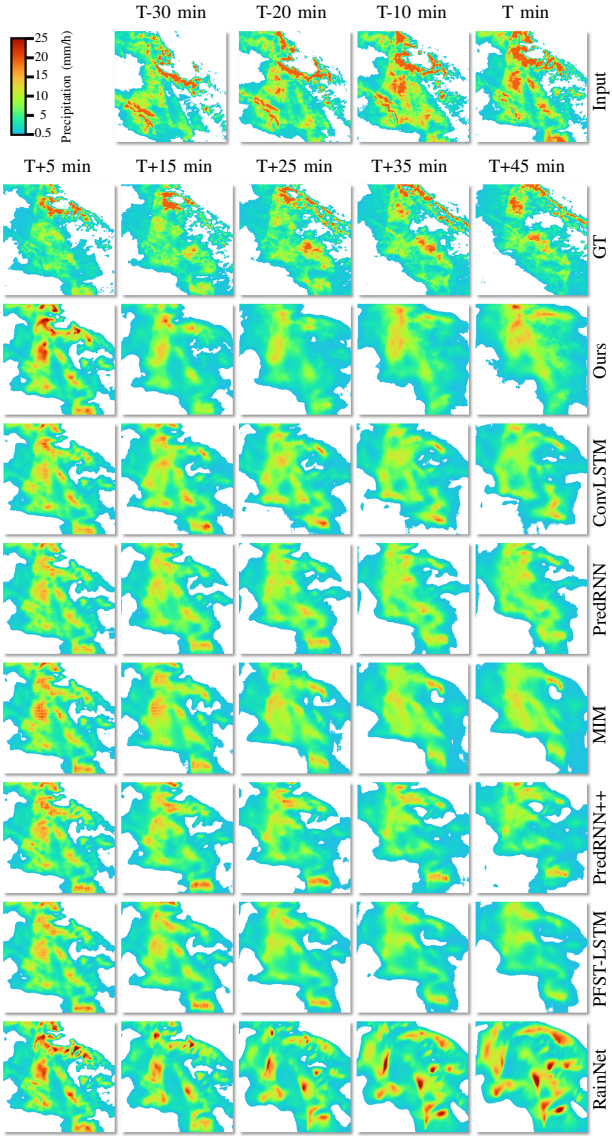

\subsubsection{Comparisons with ConvRNN based models}

Tab.~\ref{tab:quantitativeResults} also provides quantitative comparison with other ConvRNNs based methods on KNMI dataset. From this table, we can see that MMINR achieves competitive scores among all experimental results. In particular, our methods achieve the best results when the rainfall intensity $\ge2,5,10$. Due to the extreme imbalance of the dataset, light rain (i.e., rainfall intensity $\ge0.5$) has a large proportion in KNMI dataset. Therefore some light rain information is lost due to the dropout of precipitation noise, resulting in a weaker performance of our model than other models for light rain prediction. Although our model without the additional temporal information mining structure, but still achieves a close scores to those ConvRNNs based methods.

\subsubsection{Comparisons on different prediction interval with SOTAs}

We also provide all metrics curves plot with respect to different prediction interval on all methods except SmaAt-UNet, because SmaAt-UNet only predicts one frame. This comparison is shown in Fig.~\ref{fig:curve}. As can be seen in CSI and HSS plots, the proposed MMINR has better results in the early prediction interval. However, when the prediction time goes longer, the performances of all methods deteriorate. But the prediction performance of MMINR is much better than other ConvRNNs based models and also better than RainNet at rainfall intensities greater than 10 mm/h. In the B-MSE and B-MAE plots, the overall effect of MMINR is consistent with other ConvRNN methods and outperforms RainNet, because MMINR avoids the prediction errors accumulation.

\subsection{Visual Comparisons with SOTAs}

As shown in Fig.~\ref{fig:visual}, we provide a visual comparison of the proposed MMINR with other SOTAs. Our method is more accurate than other ConvRNN models in predicting the direction and the trend of rainfall movement. Besides, our model has better robustness in predicting moderate to heavy rainfall (rainfall intensity greater than 10 mm/h). Although RainNet consistently predicts heavy rainfall, most predictions are wrong, and the predicted rainfall area has severe deformation, which is mainly due to error accumulation.

\subsection{Ablation Study}
To investigate the importance of NR, we compare MMINR without NR (denoted by MMINR-NDM-SRM) and with NDM (denoted by MMINR-SRM), as shown in Tab.~\ref{tab:quantitativeResults}. From this table, we can see that MMINR-SRM outperforms MMINR-SRM-NDM in all indicators, especially at rainfall intensity above 10 mm/h, with CSI and HSS improving by 43\% ($\frac{0.0364-0.0253}{0.0253}=0.4387$) and 41\%, respectively. This also shows that NDM does reduce the model's attention to noise and improves the ability to mine semantic information.

In addition, comparing MMINR with MMINR-SRM illustrates the importance of SRM. The data show that all metrics results of MMINR are better than without SRM, and all of them have a substantial improvement, which indicates that SRM does play a complementary semantic role.

Finally, comparing MMINR with MMIR without NR (MMINR-NDM-SRM) shows the importance of the proposed NR strategy, in which MMIMR outperforms MMINR-NDM-SRM in all metrics.

\section{Conclusion}
We propose a novel Multi-frame-to-Multi-frame inference framework for precipitation nowcasting task named MMINR. It can be solved by parallel computation in the training and testing phase and avoid the accumulation of prediction errors. Furthermore, considering the precipitation noise in the radar echo maps, which absorb much attention from the deep learning model, we design a Noise Resistance strategy, which contains two modules, Noise Dropout Module and Semantic Restore Module. It reduces effectively the propagation of precipitation noise to the deep layer of the network and mitigates the problem of semantic information loss when dropout the noise. Comprehensive experimental results demonstrate that our method is efficient. For future work, we plan to add a module of temporal information acquisition to compensate for the weak ability of model to mine temporal features.






%

\bibliographystyle{IEEEtran}
\bibliography{ICPR22}



\end{document}